\documentclass[review]{elsarticle}

\usepackage{hyperref}
\usepackage[utf8]{inputenc}
\usepackage{bm}
\usepackage{amsmath}
\newcommand\norm[1]{\left\lVert#1\right\rVert}
\usepackage{algpseudocode}
\usepackage{algorithm}
\usepackage{mathtools}
\usepackage{amsmath}
\usepackage{stmaryrd}
\usepackage{amssymb}
\usepackage{multirow}
\usepackage{threeparttable}
\usepackage{array,multirow}
\usepackage{xcolor}
\usepackage{rotating}
\newcommand{\sign}{\text{sign}}
\usepackage{amsthm}
\usepackage{booktabs}
\usepackage{comment}
\usepackage{enumitem}
  \newtheoremstyle{dotless}{}{}{\itshape}{}{\bfseries}{}{ }{}
  \theoremstyle{dotless}

\journal{}

\begin{document}

\begin{frontmatter}

\title{$\ell_p$-Norm Constrained One-Class Classifier Combination}
\author[mymainaddress1]{Sepehr Nourmohammadi}
\address[mymainaddress1]{Department of Computer Engineering, Bilkent University, Ankara, Turkey}
\author[mymainaddress1]{Shervin Rahimzadeh Arashloo\corref{mycorrespondingauthor}}
\cortext[mycorrespondingauthor]
{Corresponding author}\ead{s.rahimzadeh@cs.bilkent.edu.tr}

\begin{abstract}
Classifier fusion is established as an effective methodology for boosting performance in different settings and one-class classification is no exception. In this study, we consider the one-class classifier fusion problem by modelling the sparsity/uniformity of the ensemble. To this end, we formulate a convex objective function to learn the weights in a linear ensemble model and impose a variable $\ell_{p\geq1}$-norm constraint on the weight vector. The vector-norm constraint enables the model to adapt to the intrinsic uniformity/sparsity of the ensemble in the space of base learners and acts as a (soft) classifier selection mechanism by shaping the relative magnitudes of fusion weights. Drawing on the Frank-Wolfe algorithm, we then present an effective approach to solve the formulated convex constrained optimisation problem efficiently.

We evaluate the proposed one-class classifier combination approach on multiple data sets from diverse application domains and illustrate its merits in comparison to the existing approaches.

\end{abstract}
\end{frontmatter}
\section{Introduction}
Ensemble learning, where for a learning problem multiple sources of information are fused, is known to be an effective strategy for boosting performance in various learning scenarios \cite{kittler1998combining}. Different realisations of this generic methodology may appear in accordance with the level where the fusion is practised, including data fusion, feature fusion, soft decision fusion, or hard decision fusion, {\em etc}. Classifier fusion, and in particular, a soft combination of the output scores of multiple learners has been established as a standard approach to improve classification performance in various learning scenarios \cite{kittler1998combining}. The motivating principle behind adopting a classifier fusion approach is to leverage the collective ability of multiple models, presumed to be as independent as possible, to mitigate the shortcomings of a single model, thus improving the overall performance. In general, classifier fusion approaches are expected to yield better results by - reducing the risk of selecting an inaccurate individual learner; - minimising the chances of settling for a suboptimal solution when individual learners may be stuck in local optima; - allowing for a better exploration of the potential solution space; - potentially providing a better capacity to deal with imbalanced training data; - being more capable of adapting to dynamic scenarios where the representations and labels may change over time, and - helping to mitigate the curse of dimensionality and reducing the chances of overfitting \cite{sagi2018ensemble}.

Despite its appealing properties and its widespread application in multi-class classification scenarios where significant performance improvements have been observed \cite{kittler1998combining}, the one-class classifier fusion paradigm has not been explored widely. In a one-class classification (OCC) setting, one is interested in classifying an observation as normal/positive/target or as abnormal/negative/anomaly by mainly training on positive samples \cite{RAHIMZADEHARASHLOO2022108930}. The prevalent application of OCC is often witnessed in scenarios where the accumulation of counterexamples is either highly demanding or simply infeasible \cite{krawczyk2018dynamic}, challenging binary/multi-class classification approaches. The application domain of OCC is quite diverse and spans a variety of different pattern classification problems including the detection of spoofing attacks \cite{FATEMIFAR2021107696}, detection of Deepfake videos \cite{rossler2019faceforensics++}, surveillance \cite{ZHANG2020107394}, social network applications \cite{CHAKER2017266}, {\em etc.} Depending on the available training data, the one-class learning problem may be addressed as a pure learning problem where only positive objects are used for training, or as a non-pure learning task where in addition to positive samples a number of anomalous data items also exist during the training stage.

From a general perspective, the one-class classifier combination approaches may be divided into either fixed-rule or learning-based methods. While in the fixed-rule scheme, standard predefined strategies are deployed for the fusion of base learners, in the learning-based group, the training data is utilised to derive an optimal combination of multiple one-class classifiers. Although well-studied fixed-rule classifier fusion schemes exist in the literature \cite{kittler1998combining}, they suffer from certain limitations. In particular, choosing one among multiple available fusion strategies requires a high degree of domain knowledge which may not always exist. Moreover, even in the presence of such prior domain knowledge, since any fixed fusion rule relies on certain assumptions regarding the characteristics of the underlying data and the base classifiers, even the most effective fixed-rule fusion scheme for a certain problem is not guaranteed to yield an optimal strategy. Despite these limitations, fixed-rule classifier combination schemes have been utilised to fuse one-class classifiers with varied degrees of success.

Ideally, the fusion rule is to be directly learned from the data. In practice, this challenging task may be simplified as trying to learn an optimal combination of multiple base learners, each focusing on a different aspect of the problem in a parameterised model where the parameters of the model are inferred from the training data. In the context of one-class learning, this approach has been mainly practised as a weighted sum fusion of multiple experts' scores where the weights associated with different learners are inferred from the data. In this context, although some attempts have been made towards learning-based one-class classifier fusion \cite{fatemifar2022developing} they have certain limitations. First of all, the existing learning-based one-class classifier fusion schemes typically use ad hoc objective functions to learn fusion weights. The optimality of the formulated objective functions, however, has not been investigated and is poorly understood. Furthermore, utilisation of a non-convex objective function may lead to difficulties in deriving an acceptable local optimum, in addition to the impediments in reproducibility as a result of variable local optimum. Second, the existing learning-based one-class classification methods, in the learning stage, typically use heuristic optimisation methods (such as the genetic algorithm), raising concerns regarding the derived solutions' optimality. Third, in practice, it may be advantageous to discard the least useful classifiers from the ensemble before attempting to learn fusion weights. Nevertheless, in the existing approaches, either this step is completely ignored, or in the best case scenario, heuristic schemes are followed for an initial selection of base learners \cite{FATEMIFAR20221}. Ideally, a principled mechanism, as part of the whole learning procedure, is to be used for an optimal selection of base learners.

In this study, following the common approach in the literature, we consider the one-class classifier fusion model as a weighted sum of multiple base learners' scores and try to address the limitations of the existing methods. For this purpose, we pose the learning task as a convex optimisation problem where the objective function measures the discrepancies between the expected and the obtained fused scores to minimise the classification error. We further include constraints into the learning task to capture the inherent sparsity/uniformity of the ensemble. In particular, in the proposed approach, the underlying sparsity of the ensemble is controlled via an $\ell_{p\geq1}$-norm constraint where the variable vector-norm constraint yields relatively more uniform/sparser solutions when parameter $p$ is chosen to be larger/smaller (close to $1$), respectively. The formulated constrained optimisation task is convex which facilitates efficient learning and optimisation using standard optimisation algorithms. Moreover, by virtue of convexity, the computational difficulties associated with the problem of local optima are removed. For the optimisation of the proposed constrained convex learning problem, we present an efficient method using the Frank-Wolfe algorithm with predictable accuracy.

\subsection{Contributions}
The primary contributions of this work may be summarised as follows.
\begin{itemize}
    \item We propose a new learning-based approach  for a linear fusion of multiple one-class experts;
    \item We introduce a variable $\ell_p$-norm constraint into the learning problem to facilitate controlling the uniformity/sparsity of the ensemble using only positive training objects;
    \item We extend the proposed approach to handle cases where besides the positive training sample, some anomalous objects are also available for training;
    \item We present an efficient yet effective approach for the optimisation of the proposed constrained convex learning problem with predictable accuracy using the Frank-Wolfe method \cite{pmlr-v28-jaggi13};
    \item We provide the experimental evaluation results of the proposed approach on several data sets and present a comparison between the proposed method and the existing approaches.
\end{itemize}

\subsection{Outline}
The rest of the paper is organised as follows. Section \ref{lit} reviews existing work on one-class classification fusion methods. In Section \ref{met}, our one-class classifier combination approach is presented where the proposed convex objective function and an algorithm for its optimisation are introduced. Section \ref{exp} provides the results corresponding to an evaluation of the proposed approach on multiple data sets and provides a comparison between the proposed method and other methods from the literature. Section \ref{abs} presents an ablation study to analyse the effects of different ingredients of the proposed method. Finally, Section \ref{conc} concludes the paper.

\section{Related Work}
\label{lit}
The existing approaches for one-class classifier combination may be roughly categorised as fixed-rule or learning-based where the fixed-rule methods typically follow standard classifier combination schemes \cite{kittler1998combining}. As an instance of the fixed-rule approaches, the work in \cite{10.5555/648055.744087} applies fixed-rule fusion strategies such as the sum rule for the combination of multiple one-class classifiers once their outputs are normalised. Other work \cite{10.1007/3-540-45428-4_21} stacks multiple SVDD scores into a vector and applies a cosine similarity metric for a fusion of one-class classifiers for image database retrieval. In a different study \cite{NANNI2006869}, several one-class classifiers are generated using a random subspace approach which are then combined using a max rule. The work in \cite{10.1007/978-3-540-25966-4_9} advocates the usage of different fixed-rule combination mechanisms including the sum, product, and max rules to handle missing data in classification problems. The authors in \cite{KRAWCZYK20153969} study the usefulness of one-class classifier fusion using error-correcting-output-coding (ECOC) and decision templates for the task of multi-class classification.

Learning-based methods constitute a different and important category of one-class classifier fusion techniques. As an example, the work in \cite{10.1007/978-3-540-73400-0_19} considered a weighted sum fusion for a combination of multiple fuzzy KNN classifiers where the weights were optimised using the genetic algorithm. The work in \cite{8987326} presented an OCC classifier fusion method for the face PAD problem where a weighted mean strategy to combine multiple experts' outputs is proposed. The weights are inferred using the genetic algorithm. Another study \cite{FATEMIFAR20221} proposed to combine multiple OCC learners using a weighted averaging rule by using a multi-stage pruning and optimisation approach where the weights corresponding to each learner were determined based on learning on the training data using the genetic algorithm. The authors in \cite{BERGAMINI20092117} considered a fusion of one-class SVMs for biometric applications where different combination rules including sum, weighted sum, product, min, and max were examined for fusion. The determination of weights in the weighted sum rule was realised through an exhaustive search. In another study \cite{10.1007/978-3-642-02326-2_19}, an ensemble method to fuse one-class SVMs based on bagging is proposed where samples are weighted according to their proximity to the normal class. The authors in \cite{10.1007/978-3-642-28931-6_56} proposed a combination of multiple one-class experts using the ECOC framework after selecting the candidate base learners using the genetic algorithm. The study in \cite{6977201} considers combining multiple one-class classifiers for the problem of multi-class classification. For this purpose, once the classifiers' scores are normalised, they are combined using both fixed-rule and dynamic methods where in the dynamic methods heuristic mechanisms are deployed to assign weights to different learners. The study in \cite{HE20041085} applies a decision-level fusion of multiple one-class experts where the selection of the individual experts from among a large set of candidates is performed using the genetic algorithm. The study in \cite{fatemifar2022developing} considers learning fusion weights in a weighted sum model of multiple one-class experts where the parametric weights are learned using meta-heuristic algorithms. Other work \cite{wang2020dynamic} presents a one-class ensemble system based on a dynamic ensemble model using an adaptive k-nearest neighbour rule.

In this study, in order to address the shortcomings of the existing methods, we propose a learning-based approach for a robust weighted-sum fusion of multiple one-class experts where we introduce an $\ell_{p\geq1}$-norm constraint into the learning task to facilitate controlling the uniformity/sparsity of the solution. In particular, when the optimal solution is non-uniform, the vector-norm constraint is expected to favour more competent base learners in the ensemble. We then extend the proposed method to handle cases where in addition to the positive training objects a number of anomalous training samples are also available. The proposed learning problem admits a convex formulation, and thus, circumvents the limitations of the existing methods in terms of optimality of the solution and reproducibility. Solving the formulated optimisation problem is facilitated by an effective yet efficient new approach based on the Frank-Wolfe method.

\section{Methodology}
\label{met}
Although other alternatives may exist, following the majority of the learning-based studies on one-class classifier fusion, we consider the ensemble system as a linear model of base learners, {\em i.e.} a weighted sum fusion of multiple experts. Considering $R$ one-class classifiers to be combined, the fused score for decision-making in the proposed combined system is
\begin{equation}
    h=\sum_{i=1}^{R}s_i \omega_i=\mathbf{s}^\top\boldsymbol{\omega},
\end{equation}
where the score corresponding to the $i^{th}$ classifier is denoted as $s_i$ and the corresponding weight as $\omega_i$. $\boldsymbol{\omega}$ and $\mathbf{s}$ stand for the vector collections of $\omega_i$'s and $s_i$'s, respectively while $.^\top$ denotes the transpose operation. The objective then becomes to determine $\boldsymbol{\omega}$ in a way that a minimum loss is obtained. In order to define a suitable objective function through the minimisation of which $\boldsymbol\omega$ is obtained, we define an empirical loss based on general principles of classification, discussed next.

\subsection{Objective function}
The standard approach for class assignment of a test sample is to compare its fused score ({\em i.e.} $h=\mathbf{s}^\top\boldsymbol\omega$ in this study) against a decision threshold and vote for a suitable class depending on which side of the decision threshold the fused score resides. Without loss of generality let us suppose the decision threshold is $1$ and also require all positive training samples to have a fused score bigger than $1$. In this case, a target sample with a fused score bigger than $1$ incurs no loss. We further require our loss function to linearly penalise any positive sample with a fused score smaller than $1$. These requirements lead to an empirical loss function as
\begin{equation}
    \max(0,1-\mathbf{s}^\top\boldsymbol{\omega}).
    \label{}
\end{equation}
\noindent For a data set of $n$ positive training objects we consider the cost function as the total loss over the entire data set as
\begin{equation}
\sum_{i=1}^{n} \max(0,1-\mathbf{s}_i^\top\boldsymbol{\omega}),
\end{equation}
where $\mathbf{s}_i$ denotes the vector of scores corresponding to the $i^{th}$ sample in the training set and $\boldsymbol\omega$ represents the fusion weights.

\subsection{The case of negative training samples}
\label{neg}
The objective function above applies to the pure one-class learning scenario using only target training objects. However, in a one-class learning problem, it might be the case that besides positive training objects, a few negative training samples are also accessible. Although the number of such samples might not be large, they may provide valuable information to improve the classification performance. If some negative objects are also available during the training stage, we require the fused scores of those samples to be separated from the positive training objects so as to reduce the chances of misclassification. In this case, we require the fused scores of the negative samples to be smaller than $-1$ and linearly penalise any negative sample violating this condition. Incorporating these requirements into the cost function yields
\begin{equation}
\sum_{i=1}^{n} \max(0,1-y_i\mathbf{s}_i^\top\boldsymbol{\omega}),
\end{equation}
\noindent where the class label $y_i$ is $+1$ for normal objects and $-1$ for the negative samples. One identifies the empirical loss function above as the hinge loss, used in the SVM formalism. Although prevalent in the context of SVM and related topics, hinge loss has not been previously considered for one-class classifier fusion. The advantage of the hinge loss, in this context, in addition to being convex as compared to other alternatives is that other loss functions may amplify the impact of large errors (such as the squared error loss) \cite{rousseeuw2005robust}. In contrast, the hinge loss penalises any such errors linearly. Furthermore, when the score distribution is highly skewed, such as with imbalanced data, models may be unduly influenced by contamination and outliers, skewing the model to fit these extreme cases rather than the overall trend of the normal data. In such cases, the hinge loss which creates a gap between classes may provide better separation between samples from different classes.

\subsection{Constrained learning}
In practice, one or more learners in the ensemble may provide better discriminatory capability as compared with others. Nevertheless, the existing one-class classifier fusion schemes do not address this situation in a principled way. Among the few exceptions the work in \cite{FATEMIFAR20221} tries to initially prune the base learners through a genetic algorithm optimisation stage followed by an additional round of weight optimisation. Nevertheless, the work in \cite{FATEMIFAR20221} uses separate stages for pruning and weight learning which may not lead to an optimal overall fusion scheme. Furthermore performing the optimisation with a meta-heuristic approach does not provide optimality guarantees. In principle, the pruning and weight learning stages should be merged into a single stage followed by a theoretically well-understood learning stage with also provides optimality guarantees. Motivated by the aforementioned observations, in this study, we advocate the use of a sparsity-inducing constraint which can effectively control the relative sparsity/uniformity in the domain of base learners' scores. A sparsity-inducing mechanism where the degree of sparsity is determined from the data is expected to be beneficial in handling cases where one or more base learners are more discriminative as compared with others while also helping to reduce possible over-fitting or under-fitting of the model to the data. In this context, we incorporate an $\ell_{p\geq1}$-norm constraint into the learning task to control the sparsity/uniformity of the fusion weight vector. The proposed optimisation task then reads
\begin{eqnarray}
\nonumber \min_{\boldsymbol\omega}\sum_{i=1}^{n} \max(0,1-y_i\mathbf{s}_i^\top\boldsymbol{\omega}),&&\\
\text{s.t. } \lVert\boldsymbol{\omega}\rVert_p\leq 1,&&
\label{OF}
\end{eqnarray}
\noindent where $\lVert.\rVert_p$ denotes the $\ell_p$-norm of a vector. For values of $p$ closer to $1$, the weight vector is expected to be relatively sparse while for larger values of $p$, especially for $p\rightarrow \infty$ the solution becomes more uniform. In a way, for $p$ values closer to $1$, the variable $p$-norm constraint acts as a relatively aggressive classifier selection mechanism by assigning smaller weights to less competent base learners while larger weights are assigned to those with a higher discriminatory capability. For intermediate values of $p$, the system acts as a \textit{soft} classifier selection mechanism. The optimal value for $p$ may be determined using validation data for a particular problem to optimise some performance criterion.

\subsection{Discussion}
As noted earlier, the proposed approach shares similarities with the SVM formalism \cite{VAPNIK2021108018}. More specifically, the empirical loss function in the proposed approach which is justified based on simple principles for classification, corresponds to a hinge loss adopted in the SVM formulations. Although the hinge loss has been widely used in the SVM formulations, it has not been previously utilised in the context of learning-based one-class classifier fusion. Despite this partial similarity between the proposed learning problem and that of an SVM, a distinguishing characteristic between the two (which is also an important ingredient of the proposed method) is that of a variable $\ell_{p\geq1}$-norm constraint being present in the proposed approach. While in the SVM formulations, the Euclidean norm of the discriminant is optimised, in the proposed approach, an $\ell_p$-norm of the discriminant is constrained which allows one to directly control the sparsity/uniformity of the ensemble. In the context of one-class ensemble learning, this characteristic is quite important as will be also experimentally verified in the experimental evaluation section. This fundamental difference between the proposed approach and other methods, including the SVM, is also reflected in the optimisation of the proposed objective function for which we present a new algorithm next.

\subsection{Optimisation}
Note that the objective function in Eq. \ref{OF} is convex and the constraint set for $p\geq1$ is convex too. Hence, the problem in Eq. \ref{OF} is an instance of a convex optimisation task and can be solved using convex optimisation packages such as CVX \cite{cvx}. Although generic off-the-shelf optimisation packages may be utilised to solve the constrained convex problem in Eq. \ref{OF}, they may prove to be relatively slow. As such, in this section, based on the Frank-Wolfe algorithm, we present an efficient method to solve the optimisation problem in Eq. \ref{OF}.

As an iterative method, the Frank-Wolfe technique \cite{pmlr-v28-jaggi13} has recently received increased attention because of its capability to handle constraints such as maintaining a sparse structure. The technique differs from others, including the projected gradient descent and proximal algorithms, in that it does not require a projection back onto the feasible region within every iteration. Instead, it solves a local linear approximation to the objective function within the constraint set during each cycle, and thus, naturally keeps the solution within the feasible region. In order to minimise the objective function Eq. \ref{OF}. the Frank-Wolfe algorithm iteratively updates two sets of variables $\boldsymbol{\omega}^t$ and $\boldsymbol{z}^t$ at the $t^{th}$ iteration \cite{pmlr-v28-jaggi13}:
\begin{subequations}
\begin{align}
    &\boldsymbol{z}^{t}=\operatorname*{\arg\,min}_{\lVert\boldsymbol{z}\rVert_p\leq 1}\boldsymbol{z}^\top\nabla f(\boldsymbol{\omega}^{t})\label{ssub},\\
    &\boldsymbol{\omega}^{t+1}=(1-\gamma^{t})\boldsymbol{\omega}^{t}+\gamma^{t}\boldsymbol{z}^{t},
    \label{}
\end{align}
\end{subequations}
\noindent where $\nabla f(\boldsymbol{\omega}^{t})$ denotes the gradient of the objective function in Eq. \ref{OF} w.r.t. $\boldsymbol{\omega}$ and $\gamma$ is the step size which may be computed as
\begin{align}
    \gamma^{t}=\frac{2}{2+t},
\end{align}
\noindent where $t$ stands for the iteration number. To solve the optimisation subproblem in Eq. \ref{ssub}, we form its Lagrangian as
\begin{eqnarray}
\mathcal{L} = \mathbf{z}^\top\nabla f(\boldsymbol{\omega})+\mu(\norm{\mathbf{z}}_p-1),
\end{eqnarray}
where $\mu\geq0$ is the Lagrange multiplier. It can be readily confirmed that Slater's criterion is satisfied. In this case, the KKT conditions may be written as
\begin{subequations}
\begin{align}
&\nabla_{\mathbf{z}} \mathcal{L} = \nabla f(\boldsymbol{\omega})+\mu \lVert\mathbf{z}\rVert_p^{p-1} |\mathbf{z}|^{p-1}\odot\sign(\mathbf{z})=0, \label{kkt1}\\
&\lVert\mathbf{z}\rVert_p\leq 1,\\
&\mu\geq0,\\
&\mu(\norm{\mathbf{z}}_p-1)=0,
\label{kkt4}
\end{align}
\end{subequations}
\noindent where $\odot$ stands for the element-wise multiplication. Using Eq. \ref{kkt1} we have
\begin{eqnarray}
|\mathbf{z}|^{p-1}\odot\sign(\mathbf{z})=\frac{-\nabla f(\boldsymbol{\omega})}{\mu \lVert\boldsymbol{z}\rVert_p^{p-1}}.
\label{feq}
\end{eqnarray}
As in the equation above $\mu$ appears in the denominator, in order to derive a valid solution, it must hold that $\mu\neq 0$. In this case, from Eq. \ref{kkt4}, one concludes $\norm{\mathbf{z}}_p=1$. Furthermore, by analysing the optimisation problem in Eq. \ref{ssub}, one notices that for the minimisation of the linear objective function, the signs of the elements of the optimal $\mathbf{z}$ should be opposite to those of the corresponding elements in $\nabla f(\boldsymbol\omega)$. Hence, using Eq. \ref{feq} we have
\begin{eqnarray}
\mathbf{z} = -c\hspace{0.1cm}\sign\big(\nabla f(\boldsymbol{\omega})\big)\odot|\nabla f(\boldsymbol{\omega})|^{\frac{1}{p-1}},
\label{}
\end{eqnarray}
\noindent where $c$ is a normalising constant to enforce $\rVert\mathbf{z}\lVert_p=1$. The constraint $\rVert\mathbf{z}\lVert_p=1$ may be easily satisfied by choosing $c$ as $c=\frac{1}{\lVert\nabla f(\boldsymbol{\omega})^{\frac{1}{p-1}} \rVert_p}$ to derive $\mathbf{z}$ as
\begin{eqnarray}
\mathbf{z}= \frac{-\sign(\nabla f(\boldsymbol{\omega}))\odot|\nabla f(\boldsymbol{\omega})|^{\frac{1}{p-1}}}{\lVert\nabla f(\boldsymbol{\omega})^{\frac{1}{p-1}} \rVert_p}.
\label{zeq}
\end{eqnarray}
Regarding $\nabla f(\boldsymbol{\omega})$, using Eq. \ref{OF}, one obtains
\begin{eqnarray}
\label{gradf}
\nabla f(\boldsymbol{\omega}) = \sum_{i=1}^{n}-y_i\mathbf{s}_i\llbracket 1-y_i\mathbf{s}_i^\top\boldsymbol{\omega}>0 \rrbracket,
\end{eqnarray}
\noindent where $\llbracket .\rrbracket$ stands for the Iversion brackets. The complete training procedure for the proposed approach is summarised in Algorithm \ref{alg}.

\subsubsection{The case of $p\to1^+$}
\label{pto1}
Considering the numerator in Eq. \ref{zeq} and denoting the elements of $|\nabla f(\boldsymbol{\omega})|$ as $d_1$, $d_2$, \dots, $d_R$ and its maximum entry as $d_{max}$, when $p\to1^+$ one has
\begin{flalign}
\nonumber &\lim_{p\to 1^+} |\nabla f(\boldsymbol{\omega})|^{1/(p-1)} = \lim_{p\to 1^+}\big[d_1,\dots,d_{max},\dots,d_R\big]^{1/(p-1)}\\
&=\lim_{p\to 1^+}d_{max}^{1/(p-1)}\big[(\frac{d_1}{d_{max}})^{1/(p-1)},\dots,1,\dots,(\frac{d_R}{d_{max}})^{1/(p-1)}\big].
\end{flalign}
By analysing the vector $\big[(\frac{d_1}{d_{max}})^{1/(p-1)},\dots,1,\dots,(\frac{d_R}{d_{max}})^{1/(p-1)}\big]$, one observes that except for the maximum element, all vector entries are smaller than $1$. Furthermore, we have $\lim_{p\to 1^+}1/(p-1)\to+\infty$, and hence, when $p\to1^+$, except for the maximum element, all the entries of the vector above will diminish towards $0$. Consequently, for $p\to1^+$, $\mathbf{z}$ will only possess one non-zero entry at the location of the maximum element of vector $|\nabla f(\boldsymbol{\omega})|$.

\subsubsection{The case of $p\to+\infty$}
When $p\to+\infty$, one obtains $\lim_{p\to+\infty}1/(p-1)=0$. Hence, all entries of the vector $|\nabla f(\boldsymbol{\omega})|$ in Eq. \ref{zeq} will be raised to zero, and as a result, $\mathbf{z}$ will be a uniform vector.

\begin{algorithm}[t]
\caption{$\ell_p$-norm constrained one-class classifier fusion}
\label{alg}
\begin{algorithmic}[1]
\State \textbf{Input:} score vectors $\mathbf{s}_i|_{i=1}^{n}$; labels of training samples $y_{i}|_{i=1}^{n}$; vector-norm parameter $p$; number of iterations $T$.
\State \textbf{Output:} fusion weights $\boldsymbol{\omega}$.
\State Initialisation: $\boldsymbol\omega\gets R^{-1/p}\mathbf{1}$.
\For{$t = 1, 2, \dots$, T}
\State $\nabla f(\boldsymbol{\omega}) \gets \sum_{i=1}^{n}-y_i\mathbf{s}_i\llbracket 1-y_i\mathbf{s}_i^\top\boldsymbol{\omega}>0 \rrbracket$ \Comment{gradient computation}
\State $\mathbf{z}\gets \frac{-\sign(\nabla f(\boldsymbol{\omega}))\odot|\nabla f(\boldsymbol{\omega})|^{\frac{1}{p-1}}}{\lVert\nabla f(\boldsymbol{\omega})^{\frac{1}{p-1}} \rVert_p}$ \Comment{solution of the linear subproblem}
\State $\gamma\gets\frac{2}{2+t}$ \Comment{step size update}
\State $\boldsymbol{\omega}\gets(1-\gamma)\boldsymbol{\omega}+\gamma\mathbf{z}$ \Comment{solution update}
\EndFor
\end{algorithmic}
\end{algorithm}

\subsubsection{Convergence}
It is known that the convergence of the Frank–Wolfe technique is sublinear. In other words, through the iterative process, the method yields a solution which can deviate from the true solution by at most $1/T$ once $\mathcal{O}(1/T)$ iterations are passed \cite{pmlr-v28-jaggi13}. That is, after $\mathcal{O}(1/T)$ iterations the solution $\boldsymbol\omega$ generated by the Frank-Wolfe approach satisfies $f(\boldsymbol\omega) \leq f(\boldsymbol\omega^\star)+1/T$ where $\boldsymbol\omega^\star$ denotes the true optimal solution and $f(.)$ is the objective function.

\subsection{Remarks}
As discussed above, when $p\to\infty$ the proposed approach yields a uniform weight vector which effectively corresponds to the \texttt{sum} rule for classifier fusion. On the other hand, when $p\to1^+$, only one classifier corresponding to the maximum element in $|\nabla f(\boldsymbol\omega)|$ will receive a non-zero weight. By analysing Eq. \ref{gradf}, one understands that this would be the classifier that yields the maximum sum of scores on the correct side of the threshold. Assuming that the scores of all enrolled classifiers in the ensemble are suitably normalised, the selected classifier, in this case, will represent the most confident classifier, {\em i.e.} the one with the maximum average margin. Furthermore, one recovers the conventional soft-margin linear SVM by setting $p=2$ in the proposed approach. As a result, the proposed approach yields the single best learner, the \texttt{sum} fusion rule, and the conventional SVM as special cases. That is, by varying $p$ in $[1, \infty)$, one may scan the entire spectrum of base learners, starting with the single maximum margin base classifier to the case of uniformly weighting all learners to determine the optimal strategy for the fusion. This may be realised by selecting $p$ from among multiple choices based on the characteristics of the base learners and the nature of the data to maximise some performance criterion.

\section{Experiments}
\label{exp}
The results of evaluating the proposed one-class classifier fusion approach on multiple data sets from different application domains and a comparison to other approaches from the literature are presented in this section. The rest of this section is organised as follows.
\begin{itemize}
    \item In Section \ref{deta}, the details of our implementation are provided;
    \item In Section \ref{dats}, the data sets used to evaluate the performance of the proposed approach are briefly introduced; 
    \item Section \ref{res} presents the results of evaluating the proposed approach on multiple data sets as well as a comparison to other methods.
\end{itemize}

\subsection{Implementation details}
\label{deta}
To construct a pool of one-class learners to be combined using the proposed approach, we make use of multiple standard one-class learners. Motivated by their success, the one-class classifiers we use in the current work are the Support Vector Data Description (SVDD) \cite{tax2004support}, the kernelised one-class Gaussian Process (GP) \cite{kemmler2013one}, the kernel PCA (KPCA) \cite{HOFFMANN2007863} and the one-class Gaussian Mixture Model (GMM) \cite{fatemifar2022developing}. In the SVDD, GP, and KPCA methods, we use an RBF (Gaussian) kernel function whose width is selected from $\{{0.01,0.1,0.5,1,10}\}$. The dimensionality of the subspace in the KPCA approach is selected from $\{2,6,10,\dots,n\}$ where $n$ is the number of training observations. For the one-class GMM, we fix the number of components to three and consider the minimum of the Mahalanobis distances between a test observation and the three normal components of the mixture model as the novelty score.

\sloppy In the proposed approach, the $p$ parameter that controls the sparsity/uniformity of the solution is selected from $\{32/31,16/15,8/7,4/3,2,4,8,10,100\}$. A common practice before training individual classifiers is to normalise features. Similarly, before the classifier fusion stage, scores of individual learners are normalised. In this work, we use the standard Z-score normalisation for feature normalisation. For score normalisation, motivated by its success, the recently proposed two-sided min-max score normalisation method \cite{fatemifar2022developing} is used in this study. The method sets two thresholds at the two ends of the empirical distribution of the scores so that a certain $\rho$ percentage of the scores lies outside the two thresholds and then normalises the scores using the upper and lower thresholds via min-max normalisation. In this study, $\rho$ is selected from $\{1,\dots,10\}$. We tune the parameters of the proposed method using the validation sets of each data set.

\subsection{Data sets}
\label{dats}
In the current study, we consider both generic machine learning data sets for one-class classification as well as a video-based data set for forgery detection. The ten generic machine learning datasets from UCI used in this study cover various application domains including healthcare, banknotes, detecting diseases, {\em etc}. On these data sets, the number of one-class experts in the ensemble is determined by the number of different one-class learners deployed in the current study, {\em i.e.} four. In our experiments, the generic machine learning data sets are partitioned into training, validation, and test subsets to enable a fair comparison with the state-of-the-art methods \cite{fatemifar2022developing}. For this purpose, in the pure learning scenario where only target objects are used for training, we allocate $70\%$ of normal samples to training, $20\%$ to validation, and $10\%$ to testing, while the negative data is split into $50\%$ for validation and $50\%$ for testing purposes. When operating in a non-pure scenario, half of the negative validation samples are used for training and the other half is used for validation purposes. The random division of the data set into the training, validation, and test sets is performed ten times and we report the mean and the standard deviation of the performance metrics. To enable a comparison with a wider range of the state-of-the-art methods from the literature, we report both the AUC (area under the ROC curve) as well as the G-mean defined as the square root of the product of accuracies for the positive and the negative classes.

In addition to the generic one-class classification data sets, we also include the FaceForensics++ data set \cite{rossler2019faceforensics++} in our experiments. The goal in this data set is to detect if a video depicting the face of a subject is genuine or manipulated. The data set incorporates genuine videos collected from YouTube and also manipulated video sequences generated using five different mechanisms for fake video generation, namely \texttt{FaceSwap}, \texttt{Face2Face}, \texttt{Deepfakes}, \texttt{NeuralTextures}, and \texttt{FaceShifter} for a total of more than $2.3$ million images. The data set provides video sequences corresponding to different qualities of raw, high, and low. In this work, we use the more challenging low-quality and high-quality subsets. For feature extraction on this data set, first, the face region within each frame is detected using the MTCNN approach \cite{7553523}. We then extract features from three different regions corresponding to the cropped full face, the mouth along with the surrounding area, and the nose region. Motivated by the recent work in \cite{9854878}, we make use of the filtered Fourier spectrum of each image region from which features are extracted using three deep pre-trained convolutional models of Inception-v3, DarkNet-19, and AlexNet, yielding nine sets of features. Each of the aforementioned four one-class learners is then applied to each regional feature, yielding $36$ one-class classifiers in the ensemble. Following the standard evaluation protocol for the Faceforensics++ data set \cite{rossler2019faceforensics++}, we use $72\%$ of the data for training, $14\%$ for validation, and $14\%$ for testing. On the FaceForensics++ data set, following the common approach in the literature, we report the AUC and the accuracy on each subset of the data set. A summary of the statistics of the data sets employed in this study is provided in Table \ref{datasets}. 

\begin{table}[t]
\scriptsize{
\setlength{\tabcolsep}{13pt} 
\renewcommand{\arraystretch}{1.5}
\caption{Summary of the statistics of the data sets used in the experiments.}
\centering
\begin{tabular}{c c c c c }
\hline
&Data set & Samples & Normal & Features\\
\hline
D1&Banknotes & 1372 & 762 & 4\\
D2&Ionosphere & 351 & 225 & 34\\
D3&Vote & 435 & 267 & 16\\
D4&Glass & 214 & 76 & 9\\
D5&Iris & 150 & 50 & 4\\
D6&Breast Cancer Wisconsin & 699 & 458 & 9\\
D7&Wine & 178 & 71 & 13\\
D8 & Australia & 690 & 383 & 14\\
D9 & Haberman & 306 & 225 & 3\\
D10 & Hepatitis & 155 & 123 & 19\\
D11&Faceforensics++&2,339,189&509,128&cf. \S \ref{dats}\\
\hline
\end{tabular}
\label{datasets}}
\end{table}
\begin{table}[t]
\centering
\scriptsize{
\setlength{\tabcolsep}{3pt} 
\renewcommand{\arraystretch}{1.2}
\caption{Comparison of the AUCs (mean$\pm$std\%) (upper multi-row) and G-means (mean$\pm$std\%) (lower multi-row) of the proposed approach against the single-classifier baselines and the state-of-the-art methods over 10 runs for the UCI machine learning data sets listed in Table \ref{datasets}.}
\begin{threeparttable}[t]
\resizebox{\columnwidth}{!}{
\begin{tabular}{ c c c c c c c c c c c c c c c c}
\hline
Data set & GMM & SVDD & GP & KPCA & Pure & Non-pure & Fatemifar et al.\cite{fatemifar2022developing} & DEOD\tnote{1} \cite{wang2020dynamic}  & DEOD \tnote{2}\cite{wang2020dynamic}  & FNDF\cite{wang2022robust}\\ \hline

        \multirow{2}{*}{D1} & 83.91$\pm$4.5	& 97.18$\pm$1.0 & 97.45$\pm$1.1 & 80.92$\pm$3.4 & $99.97\pm0.0$ & 99.98$\pm$0.0 & - & - & - & - \\& 98.34$\pm$0.0 & 99.17$\pm$0.2 & 90.53$\pm$0.7 & 69.23$\pm$0.2 &   $\textbf{99.80}\pm0.0$ & 99.91$\pm$ 0.0 & 99.27 & - & - & -\\
        \multirow{2}{*}{D2} & 86.07$\pm$9.3 & 90.00$\pm$5.4 & 89.39$\pm$5.0 & 85.26$\pm$8.7 & $\textbf{96.98}\pm1.0$ & 98.39$\pm$1.1 & - & 84.53 & 85.93 & 94.9\\& 62.76$\pm$1.5 & 89.08$\pm$0.9 & 90.45$\pm$0.2 & 80.58$\pm$0.2 &  $\textbf{92.93}\pm3.8$ & 93.43$\pm$ 3.1 & 90.02 & - & - & -\\
        \multirow{2}{*}{D3} & 96.67$\pm$1.4 & 78.76$\pm$1.5 & 97.55$\pm$4.3 & 67.64$\pm$1.1 &  $\textbf{98.12}\pm1.2$ & 99.99$\pm$0.0 & - & 90.76 & 92.06 & -\\& 94.05$\pm$3.4 & 77.13$\pm$5.4 & 89.37$\pm$4.7 & 67.55$\pm$3.2 &  $\textbf{95.54}\pm4.5$ & 96.54$\pm$ 3.1 & 92.57 & - & - & -\\
        \multirow{2}{*}{D4} & 86.16$\pm$6.8 & 85.13$\pm$3.0 & 88.95$\pm$4.3 & 72.23$\pm$9.1 &  $\textbf{92.39}\pm4.4$ & 95.21$\pm$3.1 & - & 83.00 & 84.34 & 62.1\\& 79.07$\pm$8.0 & 85.11$\pm$4.0 & 86.08$\pm$4.0 & 80.13$\pm$6.1 &  $90.08\pm4.0$ & 92.86$\pm$ 0.0 & - & - & - & -\\
        \multirow{2}{*}{D5} & 93.32$\pm$2.3 & 95.84$\pm$1.3 & 95.24$\pm$0.8 & 94.68$\pm$1.2 &  $\textbf{99.95}\pm0.0$ & 99.98$\pm$0.0 & - & 91.65 & 92.6 & 95.0\\& 86.02$\pm$0.0 & 90.55$\pm$0.0 & 95.91$\pm$0.0 & 96.95$\pm$0.0 & $98.99\pm0.0$ & 99.90$\pm$0.0 & - & - & - & -\\
        \multirow{2}{*}{D6} & 93.51$\pm$3.2 & 94.31$\pm$2.3 & 86.18$\pm$2.8 & 94.65$\pm$3.4 & $\textbf{98.18}\pm1.2$ & 99.39$\pm$0.4 & - & 90.77 & 90.6 & -\\& 93.50$\pm$7.4 & 78.88$\pm$4.0 & 82.36$\pm$2.1 & 81.64$\pm$3.9 &  $\textbf{97.90}\pm2.3$ & 98.20$\pm$ 2.4 & 96.54 & - & - & -\\
        \multirow{2}{*}{D7} & 85.47$\pm$5.8 & 87.22$\pm$6.3 & 87.38$\pm$7.3 & 83.96$\pm$4.0 &  $\textbf{92.83}\pm6.3$ & 94.17$\pm$4.3 & - & 81.04 & 78.90 & 90.8\\& 83.39$\pm$6.4 & 87.18$\pm$5.3 & 88.10$\pm$6.6 & 87.18$\pm$5.5 & $89.23\pm7.0$ & 90.81$\pm$ 5.1 & - & - & - & -\\
        \multirow{2}{*}{D8} & 75.64$\pm$2.1 & 73.08$\pm$2.1 & 72.57$\pm$2.1 & 66.50$\pm$2.1 & $79.73\pm0.1$ & 85.20$\pm$0.0 & - & - & - & -\\& 73.44$\pm$1.4 & 69.31$\pm$2.2 & 71.54$\pm$1.6 & 63.60$\pm$2.4 &   $\textbf{75.38}\pm0.7$ & 79.42$\pm$ 0.0 & 71.48 & - & - & -\\
         \multirow{2}{*}{D9}& 63.70$\pm$1.0 & 63.40$\pm$1.0 & 59.34$\pm$1.3 & 52.16$\pm$1.3  &  $68.29\pm0.0$ & 69.61$\pm$0.0 & - & - & - & -\\& 62.88$\pm$1.0 & 61.23$\pm$1.0 & 58.29$\pm$1.3 & 51.56$\pm$1.3 &   $\textbf{66.57}\pm0.0$ & 66.74$\pm$ 0.0 & 60.32 & - & - & -\\
         \multirow{2}{*}{D10}& 70.13$\pm$1.1 & 54.77$\pm$1.0 & 70.06$\pm$1.0 & 53.12$\pm$1.0 &  $73.43\pm0.0$ & 78.64$\pm$0.0 & - & - & - & -\\& 68.46$\pm$2.1 & 50.32$\pm$2.0 & 68.46$\pm$1.5 & 51.03$\pm$2.0 &   $\textbf{75.00}\pm0.0$ & 81.00$\pm$ 0.0 & 71.23 & - & - & -\\
\hline
\end{tabular}}
     \begin{tablenotes}
     \item[1] Adaptive KNN homogeneous.
     \item[2] Adaptive KNN heterogeneous.
   \end{tablenotes}
    \end{threeparttable}%
\label{UCI_results}}
\end{table}
\subsection{Results}
\label{res}
The results of evaluating the proposed one-class classifier fusion approach on multiple data sets along with a comparison against the baselines and the state-of-the-art methods are presented in this section.
\subsubsection{UCI data sets}
Table \ref{UCI_results} reports the performance of the proposed approach in AUC (mean$\pm$std \%) and the G-mean (mean$\pm$std \%) over ten runs on the UCI data sets listed in Table \ref{datasets}. In the table, we also report the performances corresponding to the baseline methods of GMM, SVDD, GP, the KPCA, the sum fusion of these four classifiers along other methods from the literature. The results are reported in two different settings of ``pure" and ``non-pure" where in the pure scenario only positive objects are used for training. As discussed previously in Section \ref{neg}, the proposed approach is capable of benefiting from negative samples whenever they are available during the training stage. The ``non-pure" approach in Table \ref{UCI_results} corresponds to this learning scenario where half of the negative validation data is used for training.

Considering the pure learning scenario, as may be observed from the table, the proposed approach consistently performs better than the existing methods. While on some data sets the improvement obtained by the proposed method is marginal ({\em e.g.} on D1) on some other data sets the improvements may go beyond $6\%$ in terms of the G-mean (on the D9 data set) as compared with the state-of-the-art methods. When compared to the single classifier systems utilising GMM, SVDD, GP, and KPCA it may be seen that the proposed fusion consistently improves the performance. Furthermore, the proposed $\ell_p$-norm fusion approach performs also better than the sum rule where all classifiers receive similar weights in the ensemble.

\subsubsection{FaceForensics++ data set}
In novelty detection, sometimes it may be the case that besides positive samples, a number of negative samples are also available for training but the generative process behind negative training samples may differ from that of the test set, leading to the ``unseen" novelty detection. An important example of such a learning scenario is that of face forgery detection where one tries to classify the face image/video as genuine or fake. In this section, to examine the capability of the proposed approach for ``unseen" novelty detection we use the most widely used face forgery detection data set of FaceForensics++ composed of five different types of forgeries. In the experiments on this data set, each time we train the proposed approach on four types of fake samples in addition to the positive objects, and test on the left-out forgery type. We repeat this process for each of the five types of forgeries both for the low-quality and high-quality samples of the data set. Tables \ref{cp-cross-manipulation1} and \ref{cp-cross-manipulation2} report the performances of the proposed approach in terms of the average AUC (\%) for the low- and high-quality samples and compare them with the existing approaches. As may be observed from Table \ref{cp-cross-manipulation1}, on the most challenging low-quality samples, the proposed approach performs better than the existing methods on four out of five subsets of the data set. Regarding the high-quality samples, again, the proposed approach compares very favourably with the existing approaches where on three out of five subsets, the proposed $\ell_p$-norm contained classifier fusion approach yields the state-of-the-art performance.

\begin{table}[t]
\renewcommand{\arraystretch}{1.0}
    \centering
    \scriptsize
    \caption{Comparison of different approaches on the Faceforensics++ data set in the unseen novelty detection scenario in terms of AUC (\%) for the video-level low-quality data.}
    \begin{tabular}[t]{c c c c c c }
    \toprule
        Method & \texttt{Deepfake} & \texttt{FaceSwap} & \texttt{FaceShifter} & \texttt{Face2Face} & \texttt{Neuraltextures} \\
    \midrule
        Miao et al.\cite{miao2022hierarchical} & 86.8 & - & - & 73.0 & - \\
        Sun et al.\cite{sun2021domain} & 75.6 & 68.1 & - & 72.4 & 60.8  \\
        F$^{2}$Trans-S \cite{miao2023f}& 86.9 & - & - & 76.5 & - \\
        F$^{2}$Trans-B \cite{miao2023f}& 88.7 & - & - & \textbf{77.7} & - \\
        Tran et al.\cite{tran2023learning}& 74.7 & - & - & 72.7 & - \\
        This work & \textbf{99.2} & \textbf{97.2} & \textbf{99.2} & 73.8 & \textbf{74.2} \\        
    \bottomrule
    \label{cp-cross-manipulation1}
    \end{tabular}
\end{table}
\begin{table}[t]
\renewcommand{\arraystretch}{1.0}
    \centering
    \scriptsize
    \caption{Comparison of different approaches on the Faceforensics++ data set in the unseen novelty detection scenario in terms of AUC (\%) for the video-level high-quality data.}
    \begin{tabular}[t]{c c c c c c }
    \toprule
        Method & \texttt{Deepfake} & \texttt{FaceSwap} & \texttt{FaceShifter} & \texttt{Face2Face} & \texttt{Neuraltextures} \\
    \midrule
        Sun et al.\cite{sun2021domain}&92.7 & 64.0 & - & 80.2 & 77.3  \\
        F$^{2}$Trans-S \cite{miao2023f}& 97.4 & - & - & 90.5 & - \\
        F$^{2}$Trans-B \cite{miao2023f}& 98.9 & - & - & 94.0 & \\
        Wang et al.\cite{wang2023exploiting}& 92.9 & 70.5 & - & 90.1 & 91.9 \\
        Zhu et al.\cite{zhu2023face} & 99.4 & 83.6 & - & 94.6 & 79.2 \\
        Haliassos et al.\cite{haliassos2021lips}& 99.7 & 90.1 & - & \textbf{99.7} & \textbf{99.1} \\
        This work & \textbf{99.9} & \textbf{99.9} & \textbf{99.4} & 96.2 & 97.9 \\
        
    \bottomrule
    \label{cp-cross-manipulation2}
    \end{tabular}
\end{table}

\section{Ablation study}
\label{abs}
An analysis of the impacts of different contributions offered by the proposed approach is presented in this section. To this end, we first demonstrate how changing parameter $p$ may affect the sparsity/uniformity of fusion weights and then illustrate its impact on the performance. Next, the proposed optimisation algorithm is compared against a standard convex optimisation package, namely CVX \cite{cvx} to examine its merits in terms of running time efficiency.

\subsection{Visualising the impact of varying $p$}
\begin{figure}[t]
    \centering
        \centering
        \includegraphics[scale=.19]{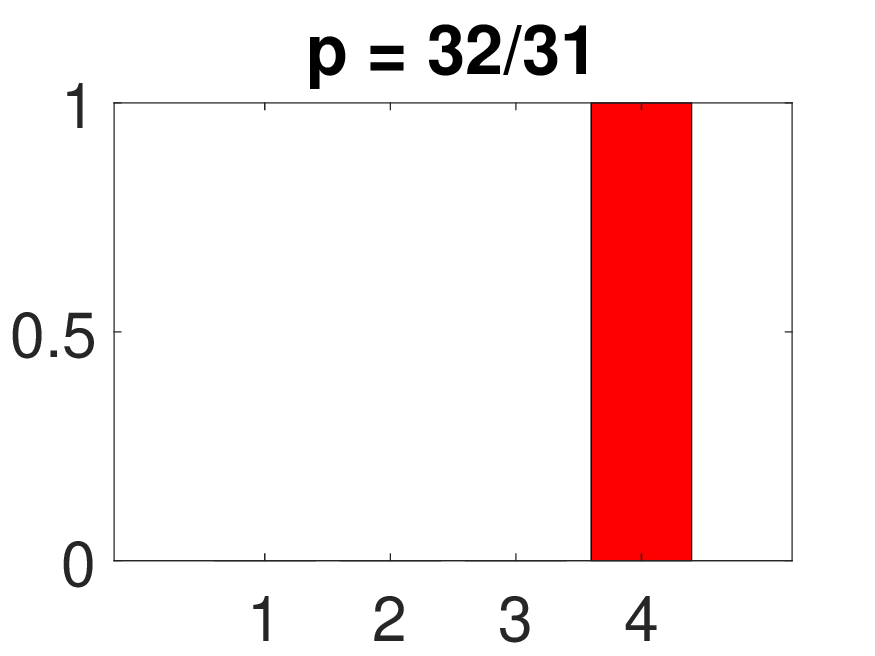}
        \includegraphics[scale=.19]{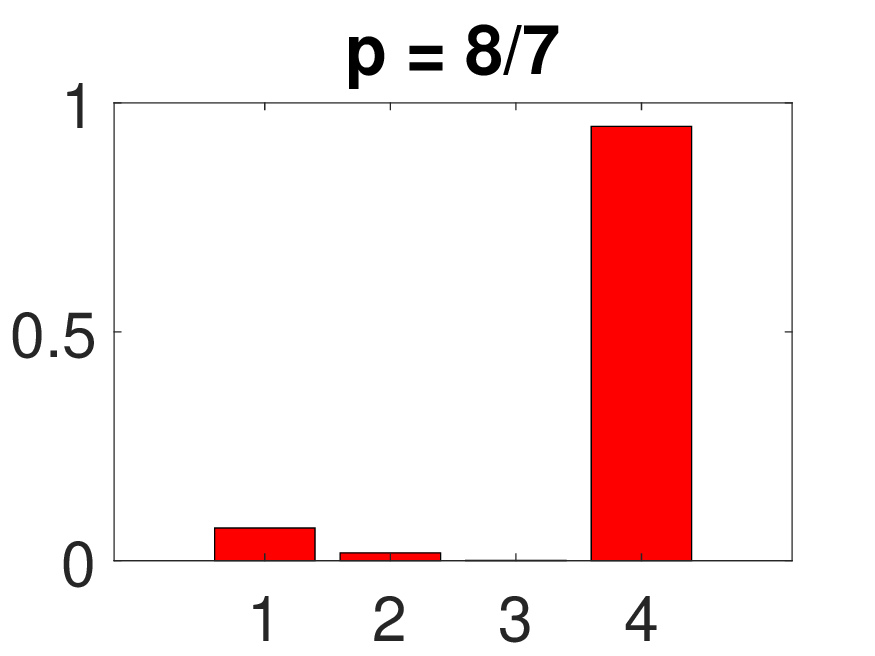}
        \includegraphics[scale=.19]{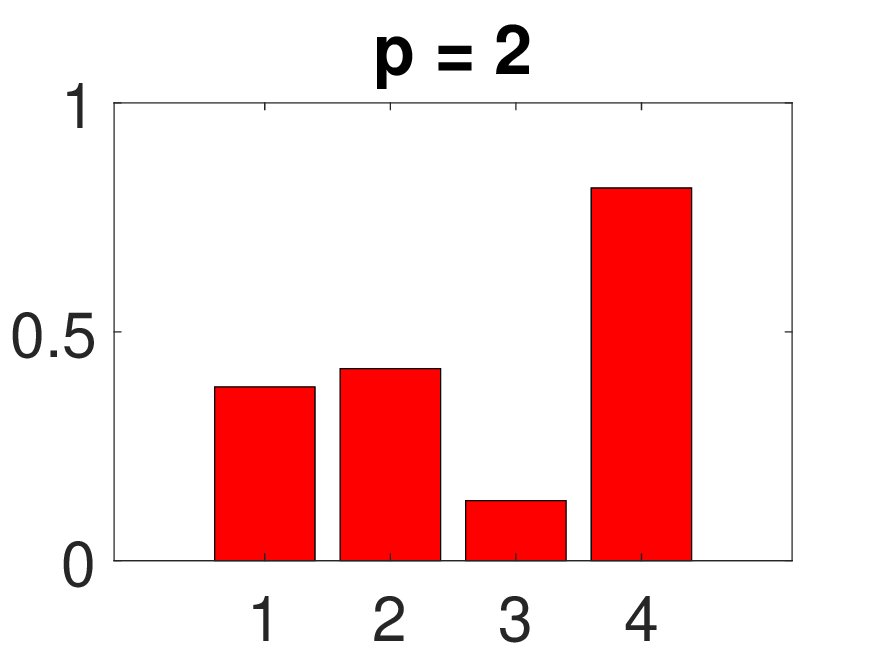}
        \includegraphics[scale=.19]{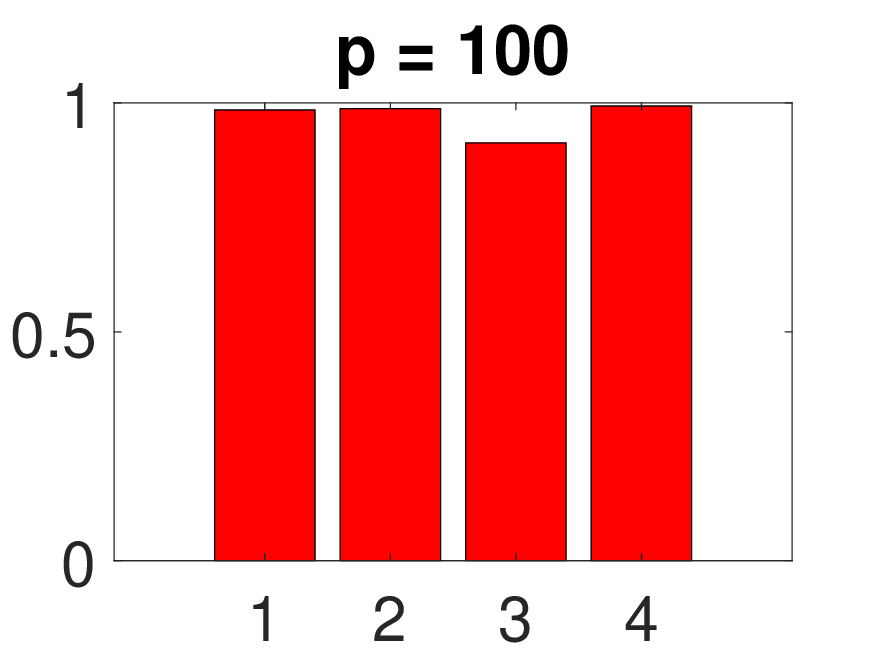}
    \caption{The effect of changing $p$ on the sparsity/uniformity of the classifier fusion weights in the proposed approach on a sample data set.}
    \label{vis}
\end{figure}

In this section, we visualise the impact of varying the norm parameter, {\em i.e.} p, on the sparsity/uniformity of the solution. For this purpose, on a sample dataset ({\em i.e.} D1: the Banknotes data set) we depict the weights inferred for four different values of $p$ selected from $\{32/31, 8/7, 2, 100\}$ in Fig. \ref{vis}. As can be inferred from the figure, for the same training data and base one-class learners, by varying $p$ different weights can be derived where by moving from $32/31$ towards $100$ relatively more uniform solutions may be obtained.

\subsection{The impact of a variable $\ell_p$-norm constraint}
An important ingredient of the proposed one-class classifier fusion approach is that of a variable $\ell_p$-norm constraint which controls the sparsity of the ensemble. In this section, we study the effect of a variable $\ell_p$-norm on the performance. For this purpose, we provide a comparison between the performance of our approach to the cases where the fusion weight vector is constrained using an $\ell_2$-norm and $\ell_\infty$-norm. Note that the case of the $\ell_2$-norm constraint resembles that of a soft-margin linear SVM while the $\ell_\infty$-norm constraint boils down to the sum fusion rule. Table \ref{lp_effect} reports the average performances of the aforementioned methods on the datasets of Table \ref{datasets}. As may be observed from the table, in the pure learning scenario where only target training samples exist, the proposed $\ell_p$-norm contained method performs better than the conventional $\ell_2$-norm as well as the sum fusion ($\ell_\infty$-norm) rule. In the non-pure learning scenario where in addition to positive samples a number of negative samples are also available for training, a similar observation can be made. While the performances of the $\ell_2$- and $\ell_\infty$-norms are very close, the proposed approach, by virtue of a variable $\ell_p$-norm constraint, performs better. A further observation from the table is that the proposed approach can effectively utilise possible negative training objects to improve the performance as the average AUC for the pure case is $89.98\%$ while that of the non-pure case is $92.05\%$.

\begin{table}[t]
\renewcommand{\arraystretch}{1.5}
    \centering
    \caption{Analysing the impact of a variable $\ell_p$-norm constraint in the proposed approach in the pure and non-pure settings in terms of average AUCs ($\%$).}
    \begin{tabular}[t]{ccccccccc}
    \hline
 &\multicolumn{4}{c}{pure learning}&\multicolumn{4}{c}{non-pure learning}\\
        \cmidrule(lr){2-4} \cmidrule(lr){5-7}
        &$\ell_2$-norm&$\ell_\infty$-norm&$\ell_p$-norm&$\ell_2$-norm&$\ell_\infty$-norm&$\ell_p$-norm\\
        \hline
         Average AUC &  87.87 & 87.90 & \textbf{89.98} &  89.13 & 89.14 & \textbf{92.05}\\
    \bottomrule
    \label{lp_effect}
    \end{tabular}
\end{table}

\subsection{The proposed Frank-Wolfe-based optimisation vs. CVX \cite{cvx}}
\begin{table}[t!]
\renewcommand{\arraystretch}{1.0}
\caption{Speed-ups achieved in the running times using the proposed Frank-Wolfe-based optimisation approach against the CVX \cite{cvx} package on a sample data set (D1) for different orders of precision and different values of $p$.}
\label{spr}
\centering
\begin{tabular}{lccc}
\hline
 & precision=1e-2& precision=1e-3&precision=1e-4\\
\hline
p = 32/31 & 898.2$\times$ & 146.0$\times$ & 134.9$\times$ \\
p = 8/7 & 19.2$\times$ & 15.9$\times$ & 14.6$\times$\\
p = 2 & 36.3$\times$ & 27.2$\times$ & 26.1$\times$\\
p = 100 & 31.0$\times$ & 21.8$\times$ & 18.8$\times$\\
\hline
\end{tabular}
\end{table}

We analyse the benefits offered by the proposed Frank-Wolfe-based optimisation scheme presented as Algorithm \ref{alg} as compared with a standard convex optimisation package, {\em i.e.} CVX \cite{cvx} in this section. For this purpose, we use different stopping conditions for the proposed iterative optimisation algorithm based on the precision defined as the maximum absolute difference between two consecutive solutions in the proposed approach. A zero value for the maximum absolute difference indicates an optimal ideal convergence. Nevertheless, in practical settings where the number of base learners in the ensemble is limited, a precision in the order of $10^{-2}$ might be sufficient. As such we examine three different orders for the precision as $10^{-2}$, $10^{-3}$, $10^{-4}$ and four different values for $p$ as $\{32/31,8/7,2,100\}$ and report the relative speed-ups obtained using the proposed approach against CVX. The results corresponding to this experiment are reported in Table \ref{spr}. As reported in the table, the proposed approach is, at least, more than $14$ times faster than the CVX \cite{cvx} package. In particular, for sparser solutions, the speed-up gain obtained using the proposed approach may reach up to $\sim 900\times$, while for more uniform solutions (when $p$ is larger) the proposed approach yields smaller speed-up gains. Yet, the proposed method as indicated above provides multiple orders of magnitude speed-up as compared with standard optimisation packages in all cases.

\subsection{The hinge loss vs. the least squares loss}
Finally, in this section, we analyse the merits of the deployed hinge loss in the context of the proposed one-class classifier fusion approach compared to other possibilities, and in particular, compared to the least squares loss function. For this experiment, we replace the hinge loss in the proposed approach with a least squares loss function and repeat the experiments on ten UCI data sets listed in Table \ref{datasets}. Table \ref{lsl}  summarises the average performance of a least squares-based approach in comparison to the hinge loss-based method in both pure and non-pure learning scenarios. As can be seen from Table \ref{lsl}, in both the pure and non-pure learning scenarios, the hinge loss-based approach yields a comparatively better performance. In particular, while in the pure learning scheme, the hinge loss-based approach provides a better average AUC by more than 1\%, for the non-pure learning approach the improvement in the average AUC is more than 2\%.
\begin{table}[t!]
\renewcommand{\arraystretch}{1.0}
\caption{Comparison of the average performance (AUC\%) of a least squares-based approach vs. the hinge loss-based method on ten data sets of Table \ref{datasets} in two different learning scenarios.}
\label{lsl}
\centering
\begin{tabular}{lcc}
\hline
 & pure learning& non-pure learning\\
\hline
Least squares-based approach &88.70&89.92\\
Hinge loss-based approach &89.98&92.05\\
\hline
\end{tabular}
\end{table}

\section{Conclusion}
\label{conc}
Considering the limitations of the existing OCC fusion methods, in this study, we introduced a new method that controls the sparsity of the ensemble through an $\ell_p$-norm constraint. Thanks to the flexibility of the vector-norm constraint for $1\leq p < \infty$, the proposed method better handles the problems associated with different discriminatory capabilities of the base classifiers in the ensemble. In particular, we demonstrated that the proposed approach recovers the conventional \texttt{sum} fusion rule, the single best learner selection, and the conventional soft-margin linear SVM as special cases. Instead of the sensitive least square loss function, we opted for the hinge loss function and formulated a constrained convex optimisation task to learn classifier fusion weights within a linear fusion scheme. Drawing on the Frank-Wolfe algorithm, we presented an efficient method to solve the proposed optimisation problem, demonstrating the practical use of our fusion framework in tasks with different dimensions. Through testing on several benchmark data sets, including multiple UCI data sets and the challenging FaceForensics++ data set, and capitalising on the varied strengths of multiple classifiers and effectively distributing weights among crucial base learners, we demonstrated that the proposed method outperforms existing approaches in the majority of cases. Although we introduced the proposed approach for pure OCC, we also presented its generalisation to the case where some negative training objects also exists which in turn paves the way for its generalisation to the multi-class setting. As potential future research directions, one may consider more flexible, {\em e.g.} non-linear fusion schemes, locally adaptive learning of the fusion model, and more effective constraints which may allow for better modelling of the inherent sparsity in the ensemble.

\section*{Acknowledgements}
This research is supported by The Scientific and Technological Research Council of Turkey (T\"{U}B\.{I}TAK) under grant no 121E465.

\bibliography{ref}

\end{document}